\documentclass[sigconf]{acmart}

\copyrightyear{2022} 
\acmYear{2022} 
\setcopyright{acmcopyright}\acmConference[MM '22]{Proceedings of the 30th ACM International Conference on Multimedia}{October 10--14, 2022}{Virtual Event, Lisbon, Portugal}
\acmBooktitle{Proceedings of the 30th ACM International Conference on Multimedia (MM '22), October 10--14, 2022, Virtual Event, Lisbon, Portugal}
\acmPrice{15.00}
\acmDOI{XX.XXXX/XXXXXXX.XXXXXXX}
\acmISBN{978-1-4503-8651-7/21/10}

\usepackage{amsmath}
\usepackage{xspace} 
\usepackage{balance}
\usepackage{makecell}
\usepackage{multirow}

\usepackage{array}
    \makeatletter
    \newcommand{\thickhline}{%
        \noalign {\ifnum 0=`}\fi \hrule height 1pt
        \futurelet \reserved@a \@xhline
}
\newcolumntype{"}{@{\vrule width 1pt}}
\makeatother

\makeatletter
\DeclareRobustCommand\onedot{\futurelet\@let@token\@onedot}
\def\@onedot{\ifx\@let@token.\else.\null\fi\xspace}

\usepackage[misc]{ifsym} 
\usepackage{graphicx}
\usepackage{relsize}

\newcolumntype{P}[1]{>{\centering\arraybackslash}p{#1}}
\makeatother

\begin{document}


\title{Exploiting Feature Diversity for Make-up Temporal \\Video Grounding}
 

\author{Xiujun Shu$^*$, Wei Wen$^*$, Taian Guo$^*$, Sunan He, Chen Wu, Ruizhi Qiao}  
\affiliation{
\institution{Tencent Youtu Lab.}
\city{}
\country{}}
\email{{xiujunshu, jawnrwen, taianguo, sunanhe, chewu, ruizhiqiao}@tencent.com}


\renewcommand{\shortauthors}{Trovato and Tobin, et al.}


\begin{abstract}
This technical report presents the 3rd winning solution for MTVG, a new task introduced in the $4^{th}$ Person in Context (PIC) Challenge at ACM MM 2022.
MTVG aims at localizing the temporal boundary of the step in an untrimmed video based on a textual description. The biggest challenge of this task is the fine-grained video-text semantics of make-up steps. However, current methods mainly extract video features using action-based pre-trained models. As actions are more coarse-grained than make-up steps, action-based features are not sufficient to provide fine-grained cues. To address this issue, we propose to achieve fine-grained representation via exploiting feature diversities. Specifically, we proposed a series of methods from feature extraction, network optimization, to model ensemble. As a result, we achieved 3rd place in the MTVG competition.

\end{abstract}




\maketitle 
\def\thefootnote{*}\footnotetext{Equal Contribution}
\section{Introduction}
Make-up Temporal Video Grounding (MTVG) is a track of the $4^{th}$ Person in Context (PIC) Challenge at ACM MM 2022. It is based on YouMakeup~\cite{wang2019youmakeup} and requires localizing the target make-up step in the video based on the textual description. In general, most video-based methods take a two-stage pipeline. First, they extract clip-level features using pre-trained models. Second, the extracted features are used for downstream tasks. We follow such a pipeline for MTVG. As most available models are pre-trained on action recognition datasets, \emph{e.g.,} Kinetics-400~\cite{carreira2017quo} and Something-Something~\cite{goyal2017something}, they are suitable for action recognition but may not be optimal for make-up scenes. This is because make-up is much more fine-grained and action-level representation may be lost some subtle cues. How to obtain fine-grained representation is critical for MTVG.

To address the above issue, we utilize diverse video modeling to increase feature diversity. This is because kinds of features could provide diverse cues from different views. It equals increasing the fine-grained cues within the representations. For the baseline framework, we have evaluated recent methods in related fields, \emph{e.g.,} temporal sentence grounding in videos (TSGV) and video moment retrieval (VMR). However, most methods, \emph{e.g.,} proposal-based~\cite{xiao2021natural}, regression-based~\cite{chen2021end}, or span-based~\cite{liu2021context}, behave much worse than 2D-Map anchor-based methods~\cite{zhang2020learning,wang2021negative}. Therefore, we choose the 2D-Map anchor-based network as our baseline framework. Besides, we find that 2D-Map splits the video into fixed number of clips, resulting in small resolution when modeling long videos. Hence, we take several methods to relieve this problem, \emph{e.g.,} multi-scale temporal sensing. Finally, we propose two strategies to ensemble the results from different models.


\section{Methodology}
In this section, we present our approach to deal with the MTVG task. 
We have three major components, \emph{i.e.,} video modeling, mutual matching network, and model ensemble. Each component will be described below in detail.


\subsection{Diverse Video Modeling} 

\subsubsection{Transformer-based Features}  
To facilitate and speed up the experiment, pre-extracted features are generally used for most video-related tasks. The competition organizer provides two features, \emph{i.e.,} C3D~\cite{tran2015learning} and I3D~\cite{carreira2017quo}. However, the experiments show us that their performances are quite different. I3D achieves 47.71\% on R1$@$0.3, but C3D only achieves 30.63\%. It tells us that features play a key role in achieving satisfying performance. In our approach, we have extracted several kinds of features using different pre-trained models. As transformer-based methods are widely used in image tasks, there are increasingly used in video tasks. ViVit~\cite{arnab2021vivit} presented four pure-transformer models for video classification, with different accuracy and efficiency. Mvit~\cite{fan2021multiscale} proposed the fundamental concept of multiscale feature hierarchies with the transformer model. video swin~\cite{liu2022video} proposed the Swin Transformer for video, which could leverage the power of the strong pre-trained image models. These pre-trained models can greatly facilitate downstream tasks.

\subsubsection{MAE-based Features}
Recently, self-supervised visual pre-training methods ~\cite{chen2021empirical}~\cite{bao2021beit} emerge, showing superior performance over supervised pre-training models.
Among them, Masked Autoencoders (MAE)~\cite{he2022masked} demonstrate superior learning ability and scalability.
Specifically, motivated by the autoencoding paradigm in BERT~\cite{devlin2018bert} in NLP, MAE adopts an asymmetric encoder-decoder architecture with visible patches encoding in the encoder and masked token reconstruction in the decoder.
VideoMAE~\cite{tong2022videomae} further extends MAE to video and shows data-efficient learners for self-supervised video pre-training.
To obtain the benefits of self-supervised learning in video classification, we utilize VideoMAE pre-trained in Something-Something~\cite{goyal2017something} to extract video features inherent in make-up scenes.

\subsubsection{Make-up Classification Features} 
As the existing pre-trained models, \emph{e.g.,} VideoSwin~\cite{liu2022video} and VideoMAE~\cite{tong2022videomae}, are pre-trained on benchmarks like action recognition, the extracted video features are inevitably more suitable to general-purpose action scenes and not optimal for the specialized-purpose make-up scenes. 
To obtain video features that are more adapted for make-up scenes, we train a make-up classification model based on the descriptions provided on the YouMakeup dataset. 
Specifically, we take the ``area'' field with 24 types in the ground truth moments descriptions as well as ``null'' as the classification categories. We optimize a VideoSwin-Base ~\cite{liu2022video} model pre-trained in Something-Something~\cite{goyal2017something} to solve the 25-class make-up area classification problem.
We speculate that although the classes for each moment are not manually annotated, the optimization objective makes the model more applicable to make-up scenes, leading to better features for temporal grounding networks.
Experiments demonstrate the advantages of make-up classification features.

\subsubsection{CLIP-based Features}
CLIP~\cite{radford2021learning} is an efficient and scalable contrastive pre-training architecture, which predicts image-text correspondence in large-scale image-text pair datasets.
Concretely, by learning on 400 million image-text pairs crawled from the Internet from scratch, CLIP learns state-of-the-art image representations, leading to amazing transfer learning capabilities on various downstream tasks.
To take advantage of the abundant prior knowledge inherent in the CLIP model pre-trained with extensive natural image-text data, we extract frame features in make-up videos with CLIP and average the frame-level features within each second to obtain make-up video features.

\begin{figure}[t]
	\centering  
	\includegraphics[width=\linewidth]{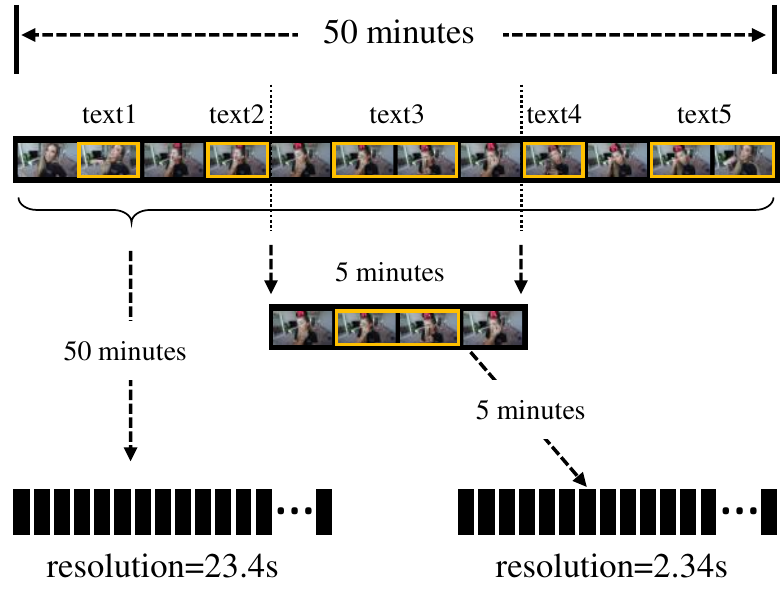}  
	\caption{Multi-scale temporal sensing.}
	\label{fig:multi_scale}
\end{figure} 

\subsection{Augmented Mutual Matching Network} 
In our approach, MMN~\cite{wang2021negative} is regarded as the baseline method. Several methods are further proposed to optimize it.

\subsubsection{Multi-scale Temporal Sensing}
2D-Map anchor-based method was first proposed in 2D-TAN~\cite{zhang2020learning}, and then was further optimized~\cite{wang2021negative,zhang2021multi}. They model the temporal relations between video moments by a two-dimensional map, which indicates the starting time and end time of moments. However, 2D-Map splits the videos into a fixed number of clips, \emph{e.g.,} 128, which results in poor temporal resolution for long videos. 

To address this issue, we propose an augmentation strategy named multi-scale temporal sensing. As shown in Fig.~\ref{fig:multi_scale}, the long video contains several textual descriptions. Each of them corresponds to specific video moments. During training, we randomly cut the video and get a clip that contains one or several complete textual descriptions. It should be noted that the start and end cutting points should be outside the textual moments to make them complete. Assume the video lasts 50 minutes and the clip length is only 5 minutes. As MMN firstly splits them into 128 pieces and each of them is pooled to get a feature. For the initial video, the temporal resolution is 23.4s (50*60/128), but the temporal resolution of the cut clip is 2.34s (5*60/128). It means the shorter the video is, the higher the temporal resolution is. By randomly clipping, the model has a stronger ability for modeling length-variable videos.

\subsubsection{Direct Feature Concatenation} 
The extracted features are not used individually, but a combination of them can achieve better performance. A simple method is the direct feature concatenation, which is described as follows:
\begin{equation}
    v = Avg(Norm(Concat(v_1, v_2, ..., v_k))),
\end{equation}
where $v_k$ is the $k^{th}$ extracted feature of the whole video. ``Concat'' denotes the feature concatenation and ``Norm'' means the $L_2$ normalization. 
``Avg'' denotes that all the features are split into 128 pieces and each of them is averaged.

Before generating the 2D-Map, the feature $v$ is firstly mapped into a small dimention with a fully connected layer $f$.
\begin{equation}
    \hat{v} = f(v).
\end{equation}

By reducing the dimension of concatenated features to a fixed value, \emph{e.g.,} 1024, the MMN structure can remain unchanged. Although the feature concatenation is simple, its performance is quite satisfactory.

\subsubsection{Learnable Feature Weighting}
Besides the concatenation, we also propose a feature weighting method in which the weighting vector is learnable.
\begin{equation}
    v_k = Avg(Norm(v_k)), k\in[1,2,3,...,K].
\end{equation}

All the K features are independently mapped into the same dimension, \emph{e.g.,} 512, then they are weighted and summed.
\begin{align}
    \hat{v} &= \gamma\cdot \sum_{k=1}^{K}\Big(\hat{w}_k\cdot v_k\Big),\\
    \hat{w}_k &= \frac{exp(w_k)}{\sum_{k=1}^{K}exp(w_k)},
\end{align}
where $w_k$ is the $k^{th}$ value of weight vector $w=[w_1,w_2,...,w_K]$. $\gamma$ is a learnable weighting parameter.

The feature concatenation and weighting operations are both utilized in our approach. They could train different models and both achieve satisfying performance.

\begin{table}[t!]
  \centering
  \small
  \renewcommand\arraystretch{1} 
  \caption{Comparison of features on YouMakeUp dataset.} 
  \begin{tabular}{p{0.5cm}|p{2.1cm}|p{0.9cm}<{\centering}|p{0.9cm}<{\centering}|p{0.9cm}<{\centering}|p{0.9cm}<{\centering}}
    \thickhline
    ID  &Feature & R1$@$0.3 & R1$@$0.5 & R1$@$0.7 & AVG \\
    \hline
    \hline
    A   &C3D             & 30.63 & 20.84 & 10.04 & 20.50 \\
    B   &I3D             & 47.71 & 33.31 & 17.49 & 32.84 \\
    C   &VideoSwin-K400  & 51.53 & 39.25 & 22.26 & 37.68 \\
    D   &VideoSwin-K600  & 56.77 & 42.50 & 23.40 & 40.89 \\
    E   &VideoSwin-SS    & 51.09 & 38.27 & 20.81 & 36.72 \\
    F   &CLIP            & 51.85 & 38.49 & 21.85 & 37.40 \\
    G   &VideoMAE        & 35.49 & 24.25 & 11.43 & 23.72 \\
    H   &SwinMC          & 56.80 & 43.13 & 24.82 & 41.58 \\
    \thickhline
  \end{tabular}\label{tab:comparison_feat}
\end{table}

\begin{table}[t!]
  \centering
  \small
  \renewcommand\arraystretch{1} 
  \caption{Experiments of feature combination. AUG: multi-scale temporal sensing. MC: Make-up classification features. FT: Fine-tune on the best model with A->H. FT$^*$: Fine-tune using the learnable feature weighting.} 
  \begin{tabular}{p{2.9cm}|p{0.9cm}<{\centering}|p{0.9cm}<{\centering}|p{0.9cm}<{\centering}|p{0.9cm}<{\centering}}

    \thickhline
    Feature & R1$@$0.3 & R1$@$0.5 & R1$@$0.7 & AVG \\
    \hline
    \hline 
    C+E             & 60.15 & 37.02 & 27.98 & 45.05 \\
    C+D+E           & 61.70 & 48.15 & 29.30 & 46.38 \\
    B+C+D+E         & 60.06 & 47.14 & 28.42 & 45.21 \\
    D+E+F+G         & 60.15 & 45.97 & 27.34 & 44.49 \\
    B+C+D+E+F+G     & 60.85 & 47.84 & 30.69 & 46.46 \\
    \hline
    B+C+D+E<-AUG    & 61.35 & 48.06 & 28.83 & 46.08 \\
    B+C+D+E+F+G+H    & 63.66 & 51.31 & 32.90 & 49.29 \\
    \hline
    B+C+D+E+F+G<-FT$^*$    & 63.91 & 50.43 & 32.18 & 48.84 \\
    B+C+D+E+F+G<-FT    & 64.22 & 51.50 & 33.03 & 49.58 \\
    \thickhline
  \end{tabular}\label{tab:comparison_feat_combine}
\end{table}

\subsection{Model Ensemble} \label{ensemble}
Based on the multiple powerful features and combination forms described above, we trained several different models with complementary information to improve the final results. Here we integrate them in two ways, \emph{i.e.,} intra-framework fusion and inter-framework fusion. 
\subsubsection{Intra-framework Fusion}
We use the same framework to train different models, in which feature combinations and parameters are different. In this case, we can do the aggregation of 2D score map inside the framework. For example, the baseline method MMN~\cite{wang2021negative} combines the contrastive map and iou map to obtain the final 2D score map, which is the key to getting the final forecast moment. 
\begin{equation}
    \textbf{S} = (S_{cons}*0.5+0.5)^{0.3}*S_{iou}
\end{equation}
Where $S_{cons}$ and $S_{iou}$ denote the contrastive map and iou map, $S$ denote the final 2D score map of the current model.

To fuse the predicted results, We aggregate 2D score maps from different models. The 2D score maps are complementary and contribute to more accurate moments.
\begin{equation} 
    \textbf{S}_{combine} = \sum_{k=1}^{n} S_{k},
\end{equation}
Where $n$ denotes the number of models, $\textbf{S}_{combine}$ denotes aggregated 2D score map.

\subsubsection{Inter-framework Fusion}
For the models from the different framework, score map sizes and types are heterogeneous and cannot be aggregated directly. We first save the candidates and scores of different models and then combine them to perform the NMS operation. It should be noted that scores of different frameworks need to be projected to the same magnitude before the NMS operation. This fusion strategy is useful because of the differences in granularity and information learned.


\noindent
\section{Experiments} 
In this section, all the experiments have been conducted on the YouMakeup dataset, which contains 1680 videos for training, 280 videos for validation, and 420 videos for testing. As the labels of the test set are not available, we report the validation results on Rank1 accuracy with three IoUs@\{0.3,0.5,0.7\}. The average value is also reported.

\subsection{Comparison of Diverse Features} 
In Table~\ref{tab:comparison_feat}, we have evaluated several features extracted from different pre-trained models. On average, one feature is extracted every 1 second. Table~\ref{tab:comparison_feat} shows us that the performance of different features varies greatly. C3D behaves the worst and only achieves 20.50\% on average Rank1. VideoSwin~\cite{liu2022video} behaves much better, especially the feature pre-trained on K600 achieves 40.89\% on average Rank1. CLIP~\cite{radford2021learning} can also achieve satisfying performance because it has been pre-trained on large-scale image-text datasets. SwinMC denotes the make-up classification feature. It achieves the best result and the average Rank1 accuracy attains to 41.58\%. This demonstrates the effectiveness of make-up classification. To increase diversity, all the features are utilized in our approach, but not independently.

\subsection{Experiments of Feature Combination} 



Based on the analysis above, we combine several features other than use them independently. In Table~\ref{tab:comparison_feat_combine}, A->H refer to the features in Table~\ref{tab:comparison_feat}. We can see that several concatenated features can achieve much better performance than single features.  For example, 
``C+D+E'' achieves 46.38\% on average Rank1, which is much better than VideoSwin-K400(37.68\%), VideoSwin-K400(40.89\%), and VideoSwin-SS(36.72\%). This is because different features bring different cues that are valuable for the MTVG task. 

We also evaluate the multi-scale temporal sensing strategy, which is short of ``AUG'' in Table~\ref{tab:comparison_feat_combine}. It improves the R1$@$0.3 from 60.06\% to 61.35\%. The make-up classification feature, \emph{i.e.,} SwinMC, can also improve the performance significantly. It increases the R1$@$0.3 from 60.85\% to 63.66\%. In addition, fine-tuning is also useful. Specifically, we initialize the model's parameters with previous best model, \emph{e.g.,} ``B+C+D+E+F+G''. Then we add more features, \emph{e.g.,} B+C+D+E+F+G+H, and fine-tune the model. In general, the fine-tuning process is quite fast and the model could achieve better performance, see FT/FT$^*$ in Table~\ref{tab:comparison_feat_combine}.


\subsection{Fusion Results}




\begin{table}[t!]
  \centering
  \small
  \renewcommand\arraystretch{1} 
  \caption{Comparison of Fusion on YouMakeUp dataset.} 
  \begin{tabular}{p{2.9cm}|p{0.9cm}<{\centering}|p{0.9cm}<{\centering}|p{0.9cm}<{\centering}|p{0.9cm}<{\centering}}

    \thickhline
    Feature & R1$@$0.3 & R1$@$0.5 & R1$@$0.7 & AVG \\
    \hline
    \hline
    intra-Fusion  & 67.54 & 56.80 & 38.39 & 54.24 \\
    (intra+inter)-Fusion  & 68.27 & 57.59 & 38.68 & 54.85 \\

    \thickhline
  \end{tabular}\label{tab:comparison_fusion}
\end{table}

In our approach, we fuse the results of 15 models. All models contain different complementary cues, we integrate their results as described in Section \ref{ensemble}. 
As shown in Table~\ref{tab:comparison_fusion}, intra-framework fusion achieves 54.24\% on average Rank1, which is improved by a large margin compared to the best single model. Limited by time, we only submitted the result of intra-framework fusion to the competition leaderboard. However, by integrating inter-framework fusion with intra-framework fusion, the performance can be further improved considerably, which demonstrated the effectiveness of our fusion strategy.


\section{Discussion}
We have attempted to evaluate the SOTA methods in the TSGV fields. However, we regret to find that many latest methods have not released their codes. For those methods that have released their codes, \emph{e.g.,} proposal-based~\cite{xiao2021natural}, regression-based~\cite{chen2021end}, and span-based~\cite{liu2021context}. They behave much worse than MMN~\cite{wang2021negative} on the YouMakeUP dataset. It may be due to the different fine-grained levels between YouMakeUP and other public datasets. This need to be further studied in the future.


\section{Conclusion}
In this report, we present our approach for MTVG, the new task introduced in the $4^{th}$ Person in Context (PIC) Challenge at ACM MM 2022. To relieve the gap between action-level and make-up features, we introduce three major components, \emph{i.e.,} diverse video modeling, augmented mutual matching network, and model ensemble. Our proposed method has achieved the 3rd place in MTVG test set.

 
 
\bibliographystyle{ACM-Reference-Format}
\balance
\bibliography{egbib} 









\end{document}